# Analyzing Emotional Trends from X platform using SenticNet: A Comparative Analysis with Cryptocurrency Price*

Moein Shahiki Tash[1], Zahra Ahani[1] Olga Kolesnikova[1], and Grigori Sidorov[1]

*Instituto Politécnico Nacional (IPN), Centro de Investigación en Computación (CIC), Mexico city, Mexico*

**Abstract**

This study delves into the relationship between emotional trends from X platform data and the market dynamics of well-known cryptocurrencies—Cardano, Binance, Fantom, Matic, and Ripple—over the period from October 2022 to March 2023. Leveraging SenticNet, we identified emotions like Fear and Anxiety, Rage and Anger, Grief and Sadness, Delight and Pleasantness, Enthusiasm and Eagerness, and Delight and Joy. Following data extraction, we segmented each month into bi-weekly intervals, replicating this process for price data obtained from Finance-Yahoo. Consequently, a comparative analysis was conducted, establishing connections between emotional trends observed across bi-weekly intervals and cryptocurrency prices, uncovering significant correlations between emotional sentiments and coin valuations.

## 1. Introduction

The influence of cryptocurrency price instability on the emotions and decision-making of investors pertains to how the rapid and unpredictable fluctuations in cryptocurrency values can affect the psychological well-being of investors and, in turn, shape their investment decisions. Starting in 2013, the cryptocurrency market has garnered considerable interest from both the media and the academic community due to its substantial price volatility [1]. The pronounced price fluctuations of cryptocurrencies have spurred an increased scholarly investigation into the origins of their value and the potential for an asset bubble. Because of its unique attributes that impact its price behavior, such as the supply and demand of Bitcoin, investor appeal, user privacy, and the involvement of computer programming enthusiasts, most academic experts assert that cryptocurrencies are disconnected from the fundamental economic and financial underpinnings [2]. The remarkable price escalation in 2017 of top cryptocurrencies like Bitcoin and Ethereum [3], has not only drawn the attention of investors but has also piqued the interest of researchers studying the operational aspects and pricing trends of cryptocurrencies [4]. In the wake of Bitcoin's success, there has been a surge in the creation of numerous cryptocurrencies. Currently, newer digital currencies like Ethereum, Ripple, Cardano, and Dogecoin are gaining substantial traction and garnering growing interest from investors [5, 6, 7, 8]. It is noteworthy that public sentiment in the digital realm, as reflected in social media, has a notable impact on the price fluctuations of cryptocurrencies [9, 10, 11].

✉ mshahikit2022@cic.ipn.mx (M. S. Tash)

In this paper, our analysis focused on five prominent cryptocurrencies, namely Ripple, Matic, Fantom, Cardano, and Binance. Our primary objective was to ascertain any correlation between cryptocurrency prices and the emotions expressed in tweets. To achieve this, we employed the SenticNet [12] model for emotion extraction, SenticNet is a freely accessible semantic and affective resource designed for tasks such as opinion mining and sentiment analysis [13] and obtained cryptocurrency price data from finance-yahoo [1]. We then calculated the percentage change in prices to determine whether they increased or decreased. The dataset of tweets we collected spanned from October 2022 to March 2023, encompassing a comprehensive six-month period. We extended our investigation by delving into the concept of "reasons and significance." Put simply, our objective was to establish a correlation between cryptocurrency prices and the emotional content expressed in tweets.

**RQ 1 :** Which cryptocurrency shows the strongest correlation between price and emotion?

**R Q 2:** Is it reliable to base trading decisions solely on emotions consistently?

**R Q 3:** Do fluctuations in price consistently correspond to shifts in positive or negative sentiment? To address this question, specific details can be found in the next sections of the paper.

The remaining content of the article is organized as follows:

- Section 2 provides an overview of previous research in the field.
- In section 3, we'll detail our methodology, encompassing the dataset, cryptocurrency price analysis, and emotional trends associated with the coins.
- Section 4 presents an in-depth analysis of the data.
- Section 5 provides the article's conclusion.

## 2. Related work

Dag et al. [14] introduce a semi-interpretable predictive framework for BTC trading decisions, employing a comprehensive feature selection approach. Building on prior research selections [15, 16, 17, 18, 19], the employed dataset focuses on factors relevant to BTC price fluctuations. The proposed Tree Augmented Naïve (TAN) model, validated through 10-fold cross-validation, achieves an average AUC of 0.652, sensitivity of 0.697, specificity of 0.641, and accuracy of 0.667, utilizing only 6 variables.

Zhou et al. [8]: in the realm of cryptocurrency portfolio optimization, this paper introduces an innovative model that integrates portfolio theory, text sentiment analysis [20], and machine learning [21] to enhance the out-of-sample performance of cryptocurrency portfolio strategies. Utilizing diverse data sources, including historical trading data, Google Trends, and X tweets, the authors employ the VADER algorithm to analyze sentiment in cryptocurrency-related tweets for predicting future price movements. Three distinct sentiment indicators are constructed, and a novel portfolio optimization model is proposed, considering both cryptocurrency forecasting results and the minimum variance portfolio model. The empirical evaluation, focusing on seven representative cryptocurrencies, demonstrates the superior out-of-sample performance of the proposed portfolio strategy compared to alternative strategies and the CRIX in most instances.

---

[1]https://finance.yahoo.com/

Valencia et al. [22]: This study showcased the feasibility of predicting price movements in the emerging cryptocurrency market using machine learning and sentiment analysis, previously applied to Bitcoin. Three prediction models—MLPs, SVMs, and RFs—were assessed for Bitcoin, Ethereum, Ripple, and Litecoin, with data sourced from X, market data, or a combination of both. Results revealed that for all four cryptocurrencies, at least one model exhibited precision scores exceeding random chance in predicting market direction. However, this predictive ability was limited to directional movements and did not encompass magnitude or duration. Notably, Litecoin emerged as the most predictable market, with the SVM model (0.66), accuracy, attaining the highest precision scores, followed by Bitcoin and Ripple.

Gurrib and Kamalov [23]: in this research, a novel methodology for predicting BTC price fluctuations was introduced, incorporating discriminant analysis and sentiment analysis. The model was subjected to training and testing using a five-year dataset encompassing daily observations up to April 2021. The data is sourced from Coinmarketcap.com. The outcomes of the experiments underscored the effectiveness of the proposed LDA approach, generating precise forecasts that surpassed random predictions. Through a comparative examination with the SVM model, it was observed that the SVM model attained the highest forecast accuracy, registering at 0.585.

Bouri and Gupta [24] conducted a study employing machine learning techniques such as ordinary least squares, support vector regression, and the least absolute selection and shrinkage operators. Their findings suggested that BTC returns exhibited weak sensitivity to trade-related uncertainties, leading to the conclusion that investors could consider cryptocurrency as a safe haven. In a distinct investigation using sentiment data derived from Google Trends to measure US-China trade tension, they employed a heterogeneous autoregressive realized volatility model. Their research revealed that uncertainty arising from US-China trade wars enhanced the forecasting accuracies of BTC returns, particularly when employing random forest classifiers.

The study conducted by Colianni et al.,[25] delving into the utilization of X data for developing strategies in cryptocurrency trading is investigated in this research. The focal point revolves around assessing the viability of utilizing supervised learning algorithms – including logistic regression, Naive Bayes, and support vector machines – post-data cleaning. The outcomes of this approach manifest in an impressive predictive accuracy, as exemplified by the final day-to-day prediction accuracy of 95% and hour-to-hour prediction accuracy of 76.23% attained by Bern.NB model.

Wołk,[26]: This study under consideration posits sentiment analysis as a computational tool capable of predicting the prices of various cryptocurrencies across distinct time intervals. Employing X and Google Trends data, the research aims to forecast short-term cryptocurrency prices, given the influence of these social media platforms on consumer decisions. Utilizing the least square linear regression (LSLR) and Bayesian ridge regression models, both integral components of the sklearn Python library, the study focuses on predicting cryptocurrency prices based on X sentiments and Google Trends data. The research shows a conclusive link between cryptocurrency price fluctuations and the influence of social media sentiment alongside web search analytics tools like Google Trends.

## 3. Methodology

### 3.1. Dataset

The dataset acquisition process was initiated by gathering X data associated with nine prominent cryptocurrencies, resulting in an extensive dataset encompassing 832,559 tweets from September 2021 to March 2023. Subsequent rigorous preprocessing led to a refined dataset of 115,899 tweets. Leveraging the Tweepy API allowed precise filtering for English content and pertinent tweets. The preprocessing steps involved URL elimination, text refinement, and detailed cryptocurrency labeling using comprehensive keyword sets. We conducted correlation analyses among these cryptocurrencies, addressing overlapping discussions systematically for accurate data categorization. For additional details regarding this dataset, further insights can be found in the accompanying research paper [27].

Following this groundwork, we identified the five most discussed coins—Cardano, Binance, Fantom, Matic, and Ripple—from the initial nine over the recent six months, detailing their respective tweet counts and associated dates. This selection was based on their prominence within the dataset. Over a span from November 2022 to March 2023, we categorized tweets for each coin, documenting the tweet counts per month in Table 1 for a comprehensive insight into their social media presence.

**Table 1**
The percentage changes in price for each month

| Date | October | | November | | December | | January | | February | | March | | Total |
|---|---|---|---|---|---|---|---|---|---|---|---|---|---|
| Coins | 1-15 | 16-31 | 1-15 | 16-30 | 1-15 | 16-31 | 1-15 | 16-31 | 1-14 | 15-28 | 1-15 | 16-31 | |
| **Cardano** | 456 | 544 | 452 | 548 | 410 | 590 | 553 | 447 | 585 | 415 | 462 | 538 | 6000 |
| **Binance** | 470 | 530 | 647 | 353 | 464 | 536 | 462 | 538 | 492 | 508 | 478 | 522 | 6000 |
| **Fantom** | 308 | 317 | 506 | 325 | 365 | 422 | 331 | 669 | 707 | 293 | 208 | 363 | 4814 |
| **Matic** | 315 | 458 | 662 | 338 | 376 | 292 | 283 | 458 | 537 | 463 | 375 | 404 | 4961 |
| **Ripple** | 612 | 388 | 420 | 580 | 483 | 517 | 471 | 529 | 471 | 529 | 408 | 592 | 6000 |

### 3.2. Price of cryptocurrency

To determine the extracted prices, we examined the closing prices of each coin over a period of two weeks, pinpointed in Table 2 with precise timestamps. This data was sourced from finance.yahoo, spanning from October 1st to March 30th. We calculated the average prices within these intervals to derive the extracted values for analysis.

**Table 2**

The percentage changes in price for each month

| Date | October | | November | | December | | January | | February | | March | |
|---|---|---|---|---|---|---|---|---|---|---|---|---|
| Coins | 1-15 | 16-31 | 1-15 | 16-30 | 1-15 | 16-31 | 1-15 | 16-31 | 1-14 | 15-28 | 1-15 | 16-31 |
| Cardano | 0.411493 | 0.377876 | 0.372258 | 0.316781 | 0.313305467 | 0.2567005 | 0.298996333 | 0.368205688 | 0.383409714 | 0.385662214 | 0.332213133 | 0.35845475 |
| BNB | 279.5512 | 284.8128 | 310.4885 | 285.2353 | 283.6924 | 244.7955 | 271.3852 | 303.5858 | 317.3984 | 309.4186 | 292.2244 | 325.1863 |
| FTM | 0.216004 | 0.214024 | 0.222298 | 0.188884 | 0.240823 | 0.202137 | 0.248134 | 0.40109 | 0.531181 | 0.502262 | 0.398447 | 0.460363 |
| Matic | 0.805863 | 0.879996 | 1.011367 | 0.855367 | 0.910583 | 0.792417 | 0.852589 | 1.041552 | 1.230901 | 1.36157 | 1.136116 | 1.123416 |
| XRP | 0.489427 | 0.464608 | 0.419536 | 0.388297 | 0.389104 | 0.350747 | 0.357821 | 0.403843 | 0.394239 | 0.387769 | 0.374125 | 0.446986 |

## 3.3. Emotions

We utilized SenticNet[2], a sentiment analysis tool, to collect emotional insights for each coin, detailed in Tables 3-7 with illustrative examples. This method enabled us to quantify emotional sentiments linked to individual coins during the specified period. Following this, we averaged emotions across various coins at two-week intervals using data columns. It's essential to highlight that tweet percentages related to emotions were part of our SenticNet-derived results. We then consolidated related emotions—such as Fear and Anxiety, Rage and Anger, Grief and Sadness, Delight and Pleasantness, Enthusiasm and Eagerness, and Delight and Joy—and calculated their averaged values over two-week periods, facilitating further analysis in our research.

### 3.3.1. Cardano

Table 3 illustrates the emotional dynamics linked to Cardano over consecutive months, segmented into two parts each. Fear and Anxiety exhibit a steady rise from October through March, peaking notably in February. Rage and Anger maintain relatively consistent levels with slight fluctuations, spiking notably in October and March. Grief and Sadness fluctuate but consistently maintain significant levels, peaking in February. Delight and Pleasantness showcase fluctuations, notably peaking in January and March. Enthusiasm and Eagerness consistently display high values, particularly peaking in November, January, and March. Delight and Joy exhibit varied fluctuations, peaking in November and March. These emotional patterns reveal shifting sentiments associated with Cardano, presenting distinct peaks in various emotions across the observed months, reflecting the evolving sentiment landscape around this cryptocurrency.

**Table 3**

The percentage changes in price for each month

| Date | October | | November | | December | | January | | February | | March | |
|---|---|---|---|---|---|---|---|---|---|---|---|---|
| Emotions | 1-15 | 16-31 | 1-15 | 16-30 | 1-15 | 16-31 | 1-15 | 16-31 | 1-14 | 15-28 | 1-15 | 16-31 |
| Fear and Anxiety | 0.009000 | 0.010789 | 0.008895 | 0.011875 | 0.014083 | 0.015624 | 0.006721 | 0.010808 | 0.008982 | 0.007080 | 0.011752 | 0.024664 |
| Rage and Anger | 0.018549 | 0.014664 | 0.017695 | 0.016827 | 0.015314 | 0.016879 | 0.009280 | 0.013311 | 0.016420 | 0.009942 | 0.010288 | 0.014788 |
| Grief and Sadness | 0.037593 | 0.027394 | 0.037465 | 0.030031 | 0.035624 | 0.039646 | 0.019228 | 0.020384 | 0.037207 | 0.027719 | 0.028032 | 0.029446 |
| Delight and Pleasantness | 0.052138 | 0.068747 | 0.062453 | 0.063472 | 0.078105 | 0.060665 | 0.095885 | 0.067201 | 0.064639 | 0.062227 | 0.069366 | 0.070201 |
| Enthusiasm and Eagerness | 0.120594 | 0.172153 | 0.155849 | 0.128311 | 0.172574 | 0.151148 | 0.136182 | 0.161275 | 0.168512 | 0.205825 | 0.181463 | 0.140561 |
| Delight and Joy | 0.037401 | 0.078937 | 0.050529 | 0.044009 | 0.045224 | 0.038252 | 0.082987 | 0.060798 | 0.048905 | 0.057402 | 0.045611 | 0.049646 |

---

[2]sentic.net/api/

### 3.3.2. Binance

Across the distinct months outlined—October through March—the emotional landscape surrounding this data exhibits noteworthy fluctuations. Fear and Anxiety display a consistent pattern, notably spiking in February and March, while Rage and Anger showcase variable levels with peaks in October and minimal values in December and January. Grief and Sadness present a varied pattern, peaking notably in October and February. Delight and Pleasantness exhibit a fluctuating trend, with peaks in November and January. Enthusiasm and Eagerness fluctuate notably, displaying higher values in December and March. Delight and Joy reveal a varying pattern across the months, peaking in October and March.

**Table 4**
The percentage changes in price for each month

| Date | October | | November | | December | | January | | February | | March | |
|---|---|---|---|---|---|---|---|---|---|---|---|---|
| Emotions | 1-15 | 16-31 | 1-15 | 16-30 | 1-15 | 16-31 | 1-15 | 16-31 | 1-14 | 15-28 | 1-15 | 16-31 |
| Fear and Anxiety | 0.051887 | 0.087807 | 0.032409 | 0.089431 | 0.120832 | 0.113492 | 0.129858 | 0.087027 | 0.097248 | 0.076943 | 0.097214 | 0.078207 |
| Rage and Anger | 0.024298 | 0.011112 | 0.019444 | 0.018587 | 0.022672 | 0.012312 | 0.006536 | 0.011159 | 0.014281 | 0.010688 | 0.011704 | 0.015816 |
| Grief and Sadness | 0.053469 | 0.046818 | 0.047327 | 0.042106 | 0.054870 | 0.037579 | 0.034651 | 0.034354 | 0.035242 | 0.033380 | 0.020891 | 0.035767 |
| Delight and Pleasantness | 0.068904 | 0.055909 | 0.033305 | 0.035658 | 0.040963 | 0.053208 | 0.063315 | 0.052512 | 0.061925 | 0.059895 | 0.046866 | 0.035767 |
| Enthusiasm and Eagerness | 0.092810 | 0.101294 | 0.177806 | 0.109303 | 0.074734 | 0.097568 | 0.092663 | 0.109198 | 0.113605 | 0.101616 | 0.094761 | 0.086078 |
| Delight and Joy | 0.053021 | 0.043969 | 0.027963 | 0.022387 | 0.038057 | 0.038727 | 0.031454 | 0.038203 | 0.041907 | 0.053018 | 0.038131 | 0.030691 |

### 3.3.3. Fantom

The data presented depicts emotional fluctuations related to the Fantom coin across bi-weekly intervals from October to March. Fear and Anxiety show sporadic peaks during the first half of November and in March's initial bi-weekly period, with minimal to no activity in other intervals. Rage and Anger exhibit varied patterns, peaking in the first half of October and January's second bi-weekly period, remaining inactive in some intervals. Grief and Sadness showcase fluctuations, peaking in the first half of November and December. Delight and Pleasantness surge notably in December's second bi-weekly period and March's second half. Enthusiasm and Eagerness soar notably in February, demonstrating heightened activity during that period. Delight and Joy depict fluctuating sentiments, peaking in February's second half and March's initial bi-weekly interval.

**Table 5**
The percentage changes in price for each month

| Date | October | | November | | December | | January | | February | | March | |
|---|---|---|---|---|---|---|---|---|---|---|---|---|
| Emotions | 1-15 | 16-31 | 1-15 | 16-30 | 1-15 | 16-31 | 1-15 | 16-31 | 1-14 | 15-28 | 1-15 | 16-31 |
| Fear and Anxiety | 0.004935 | 0.011911 | 0.006347 | 0.006312 | 0.006702 | 0 | 0.016440 | 0.005331 | 0.004226 | 0 | 0.007510 | 0.014012 |
| Rage and Anger | 0.014507 | 0.008805 | 0.008307 | 0.003165 | 0.011480 | 0 | 0.010995 | 0.012540 | 0.009705 | 0 | 0 | 0.015248 |
| Grief and Sadness | 0.015820 | 0.024893 | 0.016006 | 0.019198 | 0.014120 | 0.009636 | 0.020882 | 0.025307 | 0.009705 | 0.024974 | 0.009930 | 0.022152 |
| Delight and Pleasantness | 0.015800 | 0.027541 | 0.028451 | 0.041014 | 0.026880 | 0.021687 | 0.065117 | 0.125070 | 0.078594 | 0.059733 | 0.103581 | 0.066646 |
| Enthusiasm and Eagerness | 0.055189 | 0.055878 | 0.028451 | 0.079983 | 0.196855 | 0.155691 | 0.115604 | 0.096888 | 0.161806 | 0.099696 | 0.120841 | 0.146653 |
| Delight and Joy | 0.013011 | 0 | 0.016306 | 0.019880 | 0.015835 | 0.019582 | 0.056070 | 0.114994 | 0.068490 | 0.040939 | 0.082529 | 0.050966 |

### 3.3.4. Matic

The emotional analysis linked to Matic throughout distinct months from October to March exhibits intriguing patterns, notably indicating zero values in Fear and Anxiety during certain intervals, specifically in October and March's first half. Similarly, both Fear and Anxiety and Rage and Anger indicate minimal or no activity in the first half of October, emphasizing relatively subdued emotional expressions during those periods. However, Grief and Sadness demonstrate consistent activity across all periods, showcasing continuous emotional engagement. Delight and Pleasantness display varying levels across intervals, peaking notably in February. Enthusiasm and Eagerness exhibit substantial activity across most periods, notably peaking in October and January, with a significant drop to zero in March's second half. Delight and Joy also reveal fluctuations, with minimal activity in some intervals, particularly in March's later bi-weekly periods.

**Table 6**
The percentage changes in price for each month

| Date | October | | November | | December | | January | | February | | March | |
|---|---|---|---|---|---|---|---|---|---|---|---|---|
| Emotions | 1-15 | 16-31 | 1-15 | 16-30 | 1-15 | 16-31 | 1-15 | 16-31 | 1-14 | 15-28 | 1-15 | 16-31 |
| Fear and Anxiety | 0 | 0.010057 | 0.004157 | 0.008403 | 0.005224 | 0 | 0.005273 | 0.004875 | 0.011218 | 0.008405 | 0.006071 | 0.009741 |
| Rage and Anger | 0 | 0.007171 | 0.008014 | 0.006267 | 0.009750 | 0.007619 | 0.010399 | 0.004584 | 0.006292 | 0.005268 | 0.012049 | 0.009741 |
| Grief and Sadness | 0.011005 | 0.011504 | 0.018725 | 0.019237 | 0.012998 | 0.022690 | 0.022133 | 0.015205 | 0.020233 | 0.017105 | 0.012049 | 0.024155 |
| Delight and Pleasantness | 0.034670 | 0.064788 | 0.060461 | 0.046534 | 0.043303 | 0.047033 | 0.035444 | 0.046170 | 0.050920 | 0.053405 | 0.054925 | 0.050893 |
| Enthusiasm and Eagerness | 0.191704 | 0.161850 | 0.155364 | 0.152169 | 0.225949 | 0.136282 | 0.151560 | 0.046170 | 0.153303 | 0.156661 | 0.140920 | 0.152183 |
| Delight and Joy | 0.034695 | 0.021834 | 0.155364 | 0.016733 | 0.225949 | 0.027687 | 0.029193 | 0.022647 | 0.022881 | 0.020281 | 0.015528 | 0.018122 |

### 3.3.5. Ripple

Across the six months observed for XRP, distinct emotional patterns emerge. Fear and Anxiety depict varying levels throughout, reaching peaks in December and March's initial bi-weekly periods. Rage and Anger fluctuate notably, with peaks in December, January, and February. Grief and Sadness consistently maintain significant levels across the months, peaking notably in December and March. Delight and Pleasantness showcase peaks in January and March, indicating varying positive sentiments. Enthusiasm and Eagerness exhibit noteworthy activity, notably peaking in February and March. Delight and Joy portray varied patterns, with significant activity in February and March, reflecting the dynamic emotional shifts surrounding XRP over the observed months.

**Table 7**
The percentage changes in price for each month

| Date | October | | November | | December | | January | | February | | March | |
|---|---|---|---|---|---|---|---|---|---|---|---|---|
| Emotions | 1-15 | 16-31 | 1-15 | 16-30 | 1-15 | 16-31 | 1-15 | 16-31 | 1-14 | 15-28 | 1-15 | 16-31 |
| Fear and Anxiety | 0.008266 | 0.011120 | 0.016293 | 0.012896 | 0.009083 | 0.017865 | 0.012132 | 0.017663 | 0.018253 | 0.010413 | 0.011280 | 0.017913 |
| Rage and Anger | 0.013408 | 0.011120 | 0.021791 | 0.019890 | 0.024802 | 0.022738 | 0.016660 | 0.019453 | 0.026882 | 0.015501 | 0.021934 | 0.023101 |
| Grief and Sadness | 0.039135 | 0.046438 | 0.043332 | 0.039998 | 0.051023 | 0.045327 | 0.055123 | 0.058742 | 0.046935 | 0.037770 | 0.042467 | 0.046576 |
| Delight and Pleasantness | 0.095931 | 0.061948 | 0.082584 | 0.082414 | 0.058795 | 0.065272 | 0.055767 | 0.068765 | 0.065844 | 0.152556 | 0.129934 | 0.080083 |
| Enthusiasm and Eagerness | 0.098571 | 0.104519 | 0.087940 | 0.082413 | 0.111835 | 0.092048 | 0.092078 | 0.084087 | 0.079474 | 0.070736 | 0.129934 | 0.099245 |
| Delight and Joy | 0.041962 | 0.034688 | 0.051911 | 0.041621 | 0.041321 | 0.040418 | 0.046067 | 0.051278 | 0.038022 | 0.050334 | 0.055390 | 0.050521 |

## 4. Analysis 2

In this phase, our focus shifted towards a comparative analysis between the average emotional trends observed over the six-month period within each bi-weekly interval and the corresponding average cryptocurrency prices of five specific coins: Cardano, Binance, Fantom, Matic, and Ripple. We aimed to correlate these emotional fluctuations with the average prices of these cryptocurrencies, meticulously matching the emotional averages for every bi-week with the precise price data for the same dates. This comparison sought to unveil potential connections or patterns between emotional sentiments and cryptocurrency price movements across the specified coins throughout the observed duration. The duplicate names for the months in the chart, like two instances of October, actually represent distinct timeframes within each month. The first occurrence of each month pertains to the first bi-week, while the second instance corresponds to the second bi-week within that same month. To pinpoint the exact dates for these bi-weeks, you can refer to Table 1, where the specific dates for each bi-week within the mentioned months are detailed.

### 4.1. Cardano

Between OctSH [3] and DecSH[4], a direct correlation exists between Enthusiasm and Eagerness, aligning with the coin's price increase. However, an interesting shift occurs afterward: despite the price rising, these emotions decline. This anomaly coincides with the price stagnating until the last peak in DecFH [5]. Notably, in DecFH, the price was 0.313305, dropped to 0.256701, then increased to 0.298996. The subsequent lack of price increase until the previous high value corresponds with the decline in these emotions.

In Figure 1, Delight and Joy, as well as Enthusiasm and Eagerness, exhibit a strong correlation with each other and with the coin's price. Contrastingly, Fear and Anxiety, until DecSH, show similar fluctuations, but afterward, a gradual decrease is evident. Similar observations occur for Rage and Anger, and Grief and Sadness, where there's a correlation, but the intensity of these emotions doesn't offer substantial insights due to their lower percentages.

Moreover, Delight and Pleasantness exhibit a lack of high correlation, primarily showcasing an inverse correlation through most of the observed periods.

The examination underscores nuanced relationships between emotions (Delight and Joy, Enthusiasm and Eagerness, Delight and Pleasantness, Grief and Sadness, Fear and Anxiety, and Rage and Anger) and Cardano prices. While some emotional indices showcased direct correlations or oscillatory patterns, others exhibited inverse or stable relationships with price movements across specific bi-weekly intervals. These findings emphasize the complexity and variability in the interplay between emotional sentiment and Cardano price dynamics.

### 4.2. Binance

In Figure 2, Enthusiasm and Eagerness exhibit a clear and consistent correlation with the coin's price, often mirroring each other's movements. Similarly, Fear and Anxiety showcase a pattern of decrease alongside price increases. However, the correlation between Delight and Pleasantness appears relatively weak, evident only from January SH to March FH, aligning with periods where price and these emotions increased concurrently. Regarding other emotions like Rage and Anger, and Grief and Sadness, a strong correlation is apparent between them. Typically, one would expect these emotions to increase when the price decreases. In December, an interesting inverse pattern emerges, possibly indicating traders who initially abstained from purchasing the coin. Now, they perceive a favorable opportunity due to the coin's decreased price. This scenario presents an advantageous moment for investment. Despite the price decline, the surge in Enthusiasm and Eagerness during this month supports this interpretation.

---

[3]October Second Half
[4]December Second Half
[5]December First Half

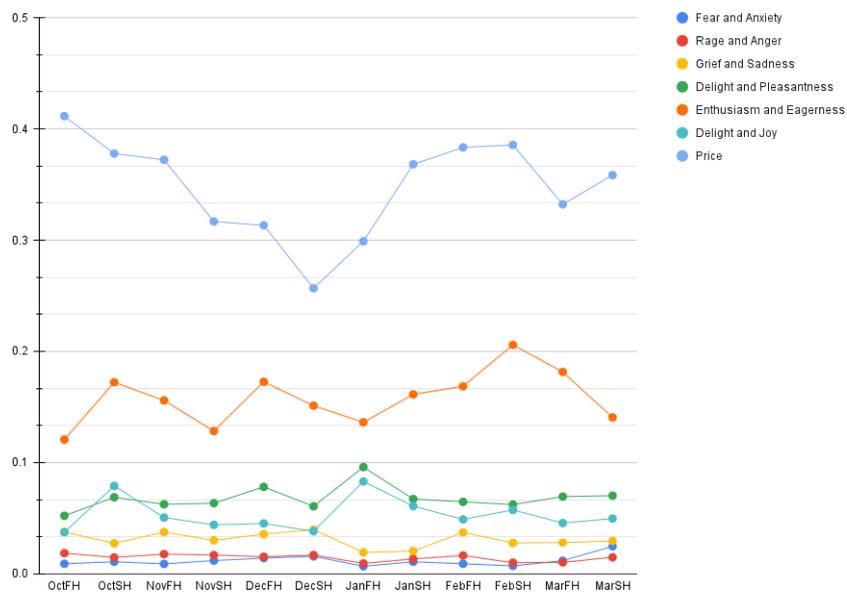

**Figure 1:** Example tweet processing

## 4.3. Fantom

In Figure 3, a notable trend emerges revealing a strong correlation between the coin's price and emotions like Enthusiasm and Eagerness across most periods. During times of price increase, these emotional factors also exhibit a noticeable uptrend. However, a deviation from this correlation occurs between DecFH and JanFH, suggesting a potential association with feelings of Fear and Anxiety. This period aligns with the sustained price stagnation, coupled with a rise in Fear and Anxiety while Enthusiasm and Eagerness decline. Subsequently, as the coin's price begins to rise again, a more typical correlation pattern reemerges.

Additionally, an inverse correlation becomes apparent between Enthusiasm and Eagerness and Fear and Anxiety. As one set of emotions rises, the other tends to decrease, indicating an interplay between these emotional states. Conversely, emotional indicators like Rage and Anger don't exhibit significant fluctuations, with values that do not provide substantial insights.

Furthermore, while other emotions display correlations, their lower values limit the insights derived from them. Delight and Pleasantness, along with Delight and Joy, notably demonstrate a strong correlation between themselves but only a weak correlation with the coin's price.

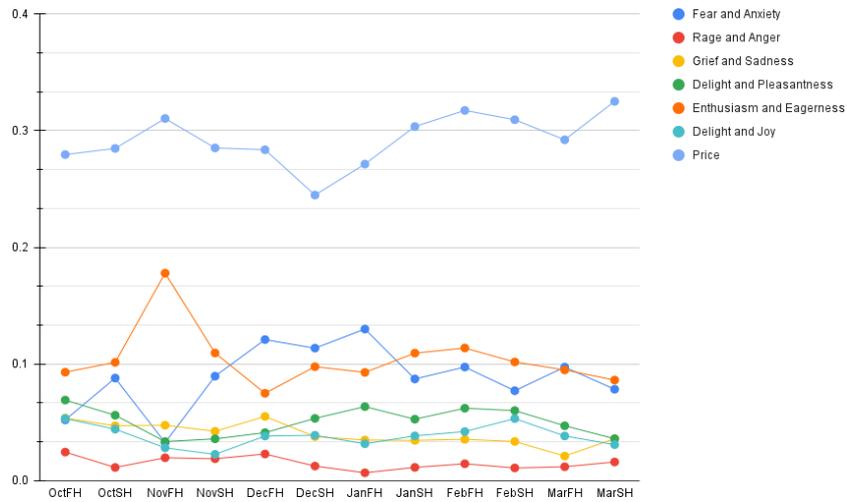

**Figure 2:** Example tweet processing

In essence, the analysis unveils a nuanced relationship between the Fantom coin's price movements and specific emotions. It highlights the significance of emotional shifts, particularly Fear and Anxiety, in influencing the coin's price trends, underscoring the potential impact of these sentiments on market dynamics.

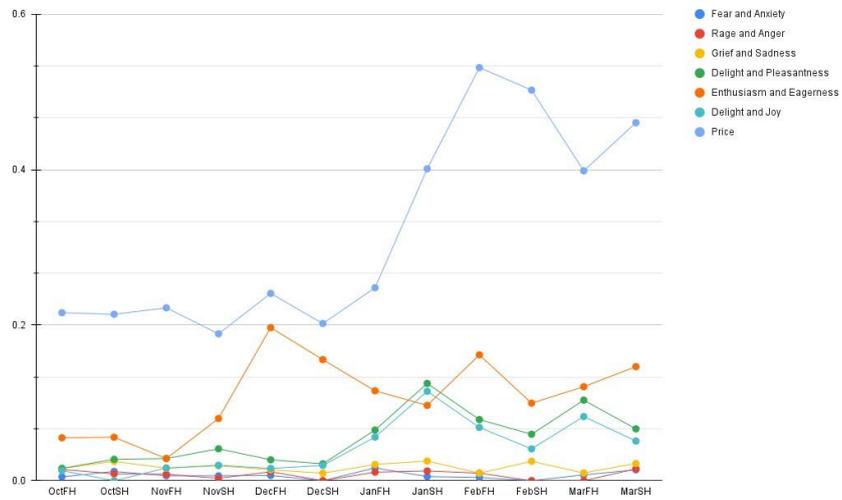

**Figure 3:** Example tweet processing

## 4.4. Matic

Analyzing Figure 4 reveals a notable pattern: the coin's price mostly saw an upward trajectory, with just one decrease observed from OctFH onward. This consistent trend toward price increase likely contributes to the low percentages of emotions like Grief and Sadness, Fear and Anxiety, and Rage and Anger. It's plausible that the infrequency of price decreases restricts the occurrences of these emotions, or this trend might be linked to the insufficiency of our dataset. In contrast, other emotions exhibit a weak correlation with the price. Notably, from DecFH onwards, when the price began to rise, there was an absence of Delight and Joy or Enthusiasm and Eagerness. This anomaly could be associated with users who miss the chance to buy the coin at a lower price and, consequently, do not express these positive emotions as the price increases significantly.

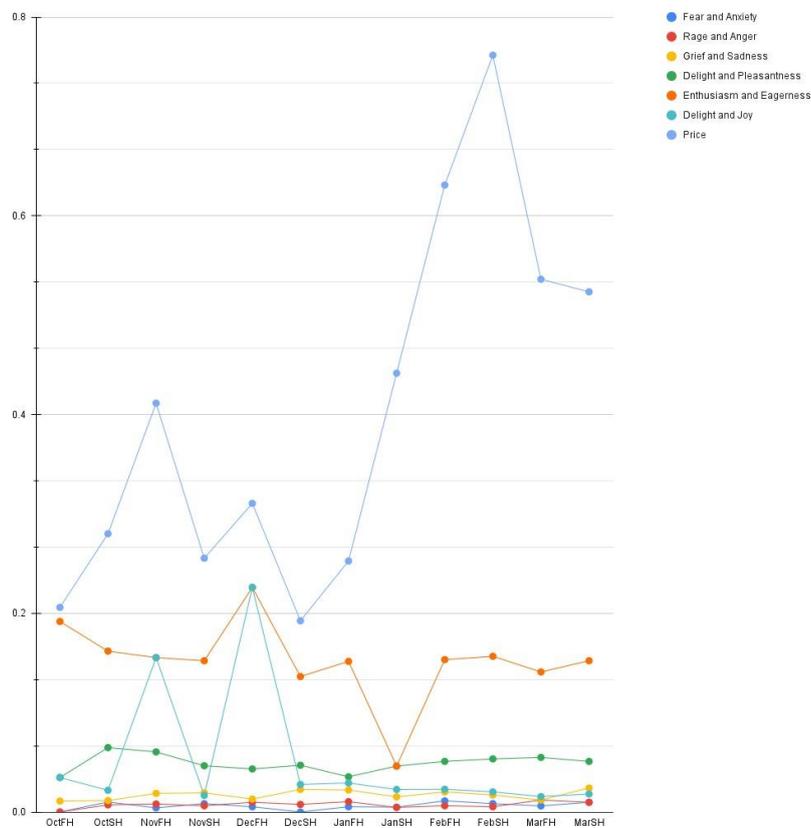

**Figure 4:** Example tweet processing

## 4.5. Ripple

Figure 5 illustrates a distinctive pattern: Fear and Anxiety consistently display an inverse swing with the price, suggesting that when the price declines, individuals may fear potential financial losses. Similar trends seem apparent for Rage and Anger and Grief and Sadness, implying a link between decreasing prices and these emotional responses. However, our dataset's limitations hinder a more nuanced analysis of these correlations. Conversely, other emotions demonstrate weak correlations, lacking any discernible special correlations among them or with the price movements. These observations underscore the complexity of emotional responses within the context of price fluctuations, highlighting the need for more extensive data to extract more conclusive insights.

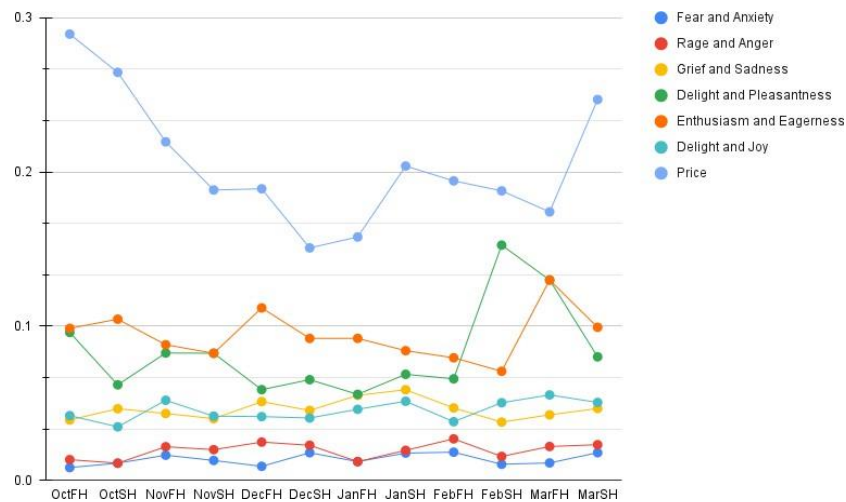

**Figure 5:** Example tweet processing

## 5. Conclusion

Based on the comprehensive comparative analysis of emotional trends and cryptocurrency prices across Cardano, Binance, Fantom, Matic, and Ripple, several intriguing connections and patterns have surfaced. In the case of Cardano, a dynamic relationship between emotional indices and price fluctuations is evident. Notably, Enthusiasm and Eagerness showcased a direct correlation with price increases, yet an unexpected decline in these emotions despite rising prices presented a unique anomaly. Delight and Joy, along with Enthusiasm and Eagerness, demonstrated substantial correlations with the coin's price, while Fear and Anxiety, Rage and Anger, and Grief and Sadness exhibited varied correlations that require deeper exploration. For Binance, a compelling alignment emerged between Enthusiasm and Eagerness and the coin's price, portraying a consistent correlation. Similarly, Fear and Anxiety mirrored price decreases, indicating a potential psychological reaction to financial uncertainties. However, other emotions displayed weaker correlations. Fantom's analysis revealed a strong correlation between price movements and emotional trends like Enthusiasm and Eagerness, emphasizing their synchronicity during periods of price increase. Notably, deviations in this correlation during phases of stagnation, coupled with a rise in Fear and Anxiety, highlighted the potential influence of emotions on market dynamics. In the case of Matic, a predominantly upward price trajectory limited the occurrences of emotions like Grief and Sadness, Fear and Anxiety, and Rage and Anger. Notably, the absence of positive emotions like Delight and Joy or Enthusiasm and Eagerness during rising prices indicated potential market sentiments among users who missed the chance to buy the coin at a lower price. For Ripple, a consistent inverse swing between Fear and Anxiety and the price suggested a correlation wherein decreasing prices induced concerns about financial losses. Similarly, trends in (Rage and Anger), (Grief and Sadness) mirrored price decreases,

In summary, our analysis consistently showed a strong correlation between positive emotions and cryptocurrency prices, notably observed in coins like Cardano and Binance, where heightened emotional swings aligned closely with price movements. However, when prices consistently decreased, as seen in XRP, or increased predominantly, like in Matic, we observed weaker correlations between emotions and price. This pattern suggests potential user segmentation: groups waiting to buy, sell, or hold coins for extended periods. Our research faced limitations, primarily stemming from limited data divided into bi-weeks, and the absence of a metric quantifying daily buy or sell percentages. To address the second research question With this information, it's clear that relying solely on emotional trends for trading decisions might not suffice; it's advisable to factor in other elements as well. For future studies, gathering more extensive data and incorporating additional metrics, such as coin volume, is essential to further unravel the intricate relationship between emotions and cryptocurrency markets.

## Acknowledgement

We express our gratitude to the SenticNet team for granting us unrestricted access to their resources.

## Funding

The work was done with partial support from the Mexican Government through the grant A1-S-47854 of CONACYT, Mexico, grants 20232138, 20231567, and 20232080 of the Secretaría de Investigación y Posgrado of the Instituto Politécnico Nacional, Mexico. The authors thank the CONACYT for the computing resources brought to them through the Plataforma de Aprendizaje Profundo para Tecnologías del Lenguaje of the Laboratorio de Supercómputo of the INAOE, Mexico and acknowledge the support of Microsoft through the Microsoft Latin America PhD Award.

## Author Contributions Statement

M.S.T., O.K., and Z.A. played pivotal roles in the experimental design and data collection, while M.S.T. and G.S. spearheaded the data analysis and interpretation. The initial manuscript was drafted by M.S.T. and O.K., with critical revisions contributed by Z.A. and G.S. All authors collectively approved the final manuscript. Notably, M.S.T., O.K., Z.A., and G.S. equally share authorship and take joint responsibility for the accuracy and integrity of the entire work.

## Conflict of Interest Statement

The authors declare that there are no conflicts of interest regarding the publication of this paper.

## Data Availability Statement

The data that support the findings of this study will be available on request from the corresponding author.